\documentclass{article}
\usepackage[utf8]{inputenc}
\usepackage{spconf,amsmath,graphicx,amssymb,amsthm,bbm}
\ninept

\usepackage[disable]{todonotes}
\usepackage{xcolor}
\usepackage{svg, import, subfiles, tikz, calc, ifthen, url, hyperref}
\usetikzlibrary{math}
\graphicspath{{Figures/}{../Figures/}}
\usetikzlibrary{decorations.pathreplacing,calligraphy}

\newcommand{\ind}[1]{^{\left(#1\right)}}
\newcommand{\vect}[1]{\mathbf{#1}}

\newcommand{\bias}{\beta}
\newcommand{\TEMLayer}[1]{\mathrm{TEM}_i\ind{#1},\, {i=1,\cdots, n\ind{#1}}}
\newcommand{\TEMLayerShort}[1]{\lbrace\mathrm{TEM}_i\ind{#1}\rbrace_i}

\def\FUND{This work was supported by the Swiss National Science Foundation grant number 200021\_181978/1, ``SESAM - Sensing and Sampling: Theory and Algorithms''.}
\name{Karen Adam\thanks{\FUND}}
\address{School of Computer and Communication Sciences\\
    Ecole Polytechnique F\'{e}d\'{e}rale de Lausanne (EPFL)}

\title{A Time Encoding approach to training Spiking Neural Networks}
\author{karen.adam }
\date{August 2021}

\begin{document}

\maketitle

\begin{abstract}
    While Spiking Neural Networks (SNNs) have been gaining in popularity, it seems that the algorithms used to train them are not powerful enough to solve the same tasks as those tackled by classical Artificial Neural Networks (ANNs).
    
    In this paper, we provide an extra tool to help us understand and train SNNs by using theory from the field of time encoding. Time encoding machines (TEMs) can be used to model integrate-and-fire neurons and have well-understood reconstruction properties.
    
    We will see how one can take inspiration from the field of TEMs to interpret the spike times of SNNs as constraints on the SNNs' weight matrices. More specifically, we study how to train one-layer SNNs by solving a set of linear constraints, and how to train two-layer SNNs by leveraging the all-or-none and asynchronous properties of the spikes emitted by SNNs.
    These properties of spikes result in an alternative to backpropagation which is not possible in the case of simultaneous and graded activations as in classical ANNs.
\end{abstract}

\section{Introduction}

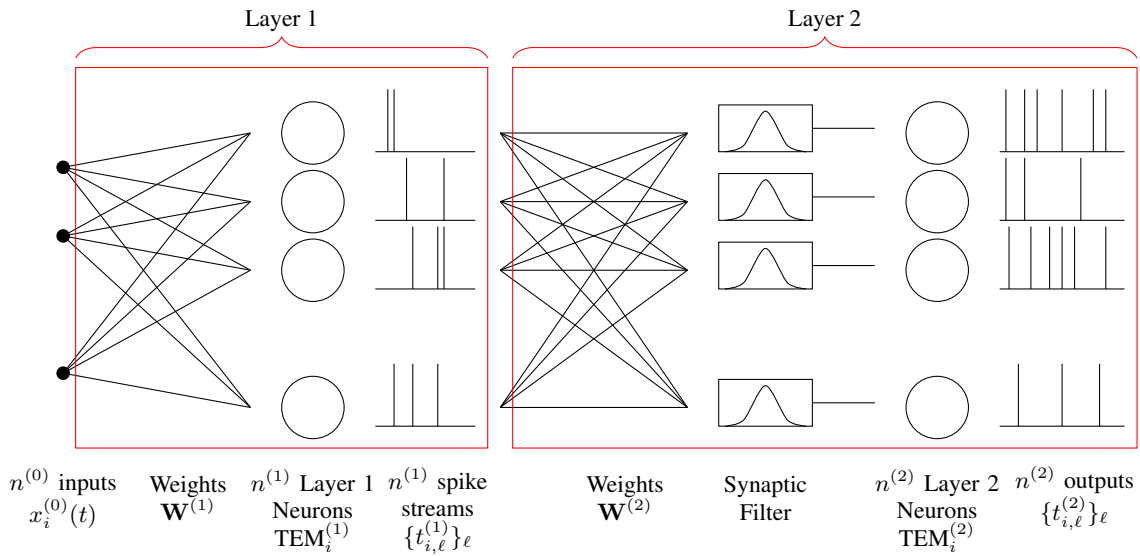
\begin{figure*}[tb]
        \centering
        \begin{center}
\begin{tikzpicture}[scale = 0.83]
\tikzmath{
    \scale = 0.83;
    \InputsX = 0;
    \TEMRectWidth=1.8;
    \NodeCircleRadius=0.5;
    \NodesXPos=0;
    \TEMRectHeight=0.9;
    \CenterX=1;
    \TEMStartCoordY=1;
    \TEMVertSpacing=0.2;
    \dotdotdotsize=0.025;
    \dotdotdotspace=0.1;
    \yToTEMLineLength=0.5;
    \XtoYLineXSpan=3;
    \YHorSpace=1.5;
    \TEMOutputLineSpan=1.5;
    \TEMSpikeLineSpan=2;
    \TEMOutSpikeVertSpace=0.3;
    \TEMOutSpikeHorSpace=0.5;
    \SpikeHeight=1;
    \FilterSpacing = 2;
    \FilterWidth=1.5;
    \FilterHeight=0.75;
    \FilterXPos=1.5;
    \FilterToTEMLineWidth=1;
}
\foreach \n in {0,1,2,3}
{
    
    \tikzmath{
        \starty = \TEMStartCoordY - \n*(\TEMRectHeight+\TEMVertSpacing);
    }
    \ifnum \n = 3
        \tikzmath{
            \starty = \TEMStartCoordY - 4*(\TEMRectHeight+\TEMVertSpacing);
        }
    \fi
    \tikzmath{
        \linestartX = (\CenterX - \yToTEMLineLength);
        \lineY = \starty+\TEMRectHeight/2;
        \startSpikey = \lineY - \TEMOutSpikeVertSpace;
    }
    \draw (\NodesXPos - \XtoYLineXSpan - 4*\NodeCircleRadius, \lineY) circle (\NodeCircleRadius);

    \draw (\NodesXPos + \TEMSpikeLineSpan + \FilterSpacing + \FilterToTEMLineWidth, \lineY) circle (\NodeCircleRadius);
    \draw  (\FilterXPos, \lineY - 0.3) rectangle(\FilterXPos + \FilterWidth , \lineY +\FilterHeight - 0.3);
    
    \draw plot [smooth] coordinates {(\FilterXPos + 0.1, \lineY - 0.3) (\FilterXPos + 0.4, \lineY - 0.2) (\FilterXPos +\FilterWidth/2, \lineY - 0.4 + \FilterHeight)(\FilterXPos +\FilterWidth- 0.4, \lineY - 0.2)(\FilterXPos +\FilterWidth - 0.1, \lineY - 0.3) };
    \ifnum \n = 0
        \foreach \loc in {0.1,0.4,0.6,1,1.5, 1.7}
        {
            \draw (\CenterX + \loc+ \TEMSpikeLineSpan+ \FilterSpacing+ \FilterToTEMLineWidth, \startSpikey) -- (\CenterX + \loc+ \TEMSpikeLineSpan+ \FilterSpacing+ \FilterToTEMLineWidth, \startSpikey + \SpikeHeight);
        }
        \foreach \loc in {0.2, 0.3}
        {
            \draw (\CenterX + \loc- \XtoYLineXSpan- 4*\NodeCircleRadius, \startSpikey) -- (\CenterX + \loc- 4*\NodeCircleRadius- \XtoYLineXSpan, \startSpikey + \SpikeHeight);

        }
    \fi
    \ifnum \n = 1
        \foreach \loc in {0.1, 0.4, 1.3}
        {
            \draw (\CenterX + \loc+ \TEMSpikeLineSpan+ \FilterSpacing+ \FilterToTEMLineWidth, \startSpikey) -- (\CenterX + \loc+ \TEMSpikeLineSpan+ \FilterSpacing+ \FilterToTEMLineWidth, \startSpikey + \SpikeHeight);
        }
        \foreach \loc in {0.5, 1.1}
        {
            \draw (\CenterX + \loc- 4*\NodeCircleRadius- \XtoYLineXSpan, \startSpikey) -- (\CenterX + \loc- 4*\NodeCircleRadius- \XtoYLineXSpan, \startSpikey + \SpikeHeight);

        }
    \fi
    \ifnum \n = 2
        \foreach \loc in {0.15,0.5,0.8,1,1.2,1.7}
        {
            \draw (\CenterX + \loc+ \TEMSpikeLineSpan+ \FilterSpacing+ \FilterToTEMLineWidth, \startSpikey) -- (\CenterX + \loc+ \TEMSpikeLineSpan+ \FilterSpacing+ \FilterToTEMLineWidth, \startSpikey + \SpikeHeight);
        }
        \foreach \loc in {0.6, 1.0, 1.1}
        {
            \draw (\CenterX + \loc- 4*\NodeCircleRadius- \XtoYLineXSpan, \startSpikey) -- (\CenterX + \loc- 4*\NodeCircleRadius- \XtoYLineXSpan, \startSpikey + \SpikeHeight);

        }
    \fi
    \ifnum \n = 3
        \foreach \loc in {0.3,1, 1.6}
        {
            \draw (\CenterX + \loc+ \TEMSpikeLineSpan+ \FilterSpacing+ \FilterToTEMLineWidth, \startSpikey) -- (\CenterX + \loc+ \TEMSpikeLineSpan+ \FilterSpacing+ \FilterToTEMLineWidth, \startSpikey + \SpikeHeight);
        }
        \foreach \loc in {0.3, 0.6,1}
        {
            \draw (\CenterX + \loc- 4*\NodeCircleRadius- \XtoYLineXSpan, \startSpikey) -- (\CenterX - 4*\NodeCircleRadius+ \loc- \XtoYLineXSpan, \startSpikey + \SpikeHeight);
        }
    \fi
    
    \draw (\CenterX+ \TEMSpikeLineSpan+ \FilterSpacing+ \FilterToTEMLineWidth, \startSpikey) -- (\CenterX+ \TEMSpikeLineSpan + \TEMSpikeLineSpan+ \FilterSpacing+ \FilterToTEMLineWidth, \startSpikey);
    \draw (\CenterX- 4*\NodeCircleRadius- \XtoYLineXSpan, \startSpikey) -- (\CenterX - 4*\NodeCircleRadius- \XtoYLineXSpan+ \TEMSpikeLineSpan*0.8, \startSpikey);
    
    \draw (\FilterXPos + \FilterWidth , \lineY +0.5*\FilterHeight - 0.3) -- (\FilterXPos + \FilterWidth +\FilterToTEMLineWidth , \lineY +0.5*\FilterHeight - 0.3);

}

\foreach \n in {0,1,2,3}
{
    \tikzmath{
        \lineStartX = \CenterX - \yToTEMLineLength - \XtoYLineXSpan - \YHorSpace + \TEMSpikeLineSpan;
    }
        \ifnum \n < 3
        \tikzmath{\lineStartY = \TEMStartCoordY - \n*1.1 + \TEMRectHeight/2;}

        \else
        \tikzmath{\lineStartY = \TEMStartCoordY - (\n+1)*(\TEMRectHeight+\TEMVertSpacing)+ \TEMRectHeight/2;}
        \fi

    \foreach \m in {0,1,2,3}
    {
        \tikzmath{\lineEndX = \CenterX - \yToTEMLineLength -\YHorSpace + + \TEMSpikeLineSpan;}
        \ifnum \m < 3
        \tikzmath{\lineEndY = \TEMStartCoordY - \m*(\TEMRectHeight+\TEMVertSpacing) +\TEMRectHeight/2;}
        \else
        \tikzmath{\lineEndY = \TEMStartCoordY - (\m+1)*(\TEMRectHeight+\TEMVertSpacing) +\TEMRectHeight/2;}
        \fi
    
            \draw (\lineStartX, \lineStartY) -- (\lineEndX, \lineEndY);
            
    }
}

\foreach \n in {0,1,2}
{
    \tikzmath{
        \lineStartX = \CenterX - \yToTEMLineLength - \XtoYLineXSpan - \YHorSpace;
    }
        \ifnum \n < 2
        \tikzmath{\lineStartY = \TEMStartCoordY - \n*(\TEMRectHeight+\TEMVertSpacing) - \TEMVertSpacing/2;}
        \else
        \tikzmath{\lineStartY = \TEMStartCoordY - (\n+1)*(\TEMRectHeight+\TEMVertSpacing) - \TEMVertSpacing/2;}
        \fi

    \foreach \m in {0,1,2,3}
    {
        \tikzmath{\lineEndX = \CenterX - \yToTEMLineLength -\YHorSpace;}
        \ifnum \m < 3
        \tikzmath{\lineEndY = \TEMStartCoordY - \m*(\TEMRectHeight+\TEMVertSpacing) +\TEMRectHeight/2;}
        \else
        \tikzmath{\lineEndY = \TEMStartCoordY - (\m+1)*(\TEMRectHeight+\TEMVertSpacing) +\TEMRectHeight/2;}
        \fi
    
            \draw (\lineStartX- \XtoYLineXSpan - 4*\NodeCircleRadius, \lineStartY) -- (\lineEndX- \XtoYLineXSpan - 4*\NodeCircleRadius, \lineEndY);
            
    }
    \filldraw (\lineStartX- \XtoYLineXSpan - 4*\NodeCircleRadius, \lineStartY) circle (0.1);
    \ifnum \n = 2
        \draw (\lineStartX - \XtoYLineXSpan - 4*\NodeCircleRadius, \lineStartY -2*\TEMRectHeight -0.2) node[text width=50,align=center]{$n\ind{0}$ inputs\\ $x_i\ind{0}(t)$};
        \draw (0.5*\lineStartX  +1.5  - \XtoYLineXSpan - 7*\NodeCircleRadius, \lineStartY -2*\TEMRectHeight-0.2) node[text width=40,align=center]{Weights $\vect{W}\ind{1}$};
        \draw (0.5*\lineStartX  -1 + 1.5*\TEMSpikeLineSpan, \lineStartY -2*\TEMRectHeight-0.2) node[text width=40,align=center]{Weights $\vect{W}\ind{2}$};

        \draw (\NodesXPos+ \TEMSpikeLineSpan + \FilterSpacing + \FilterToTEMLineWidth, \lineStartY -2*\TEMRectHeight - 0.4) node[text width=60,align=center]{$n\ind{2}$ Layer 2\\ Neurons\\ TEM$_i\ind{2}$};
        \draw (\NodesXPos - \XtoYLineXSpan - 4*\NodeCircleRadius, \lineStartY -2*\TEMRectHeight -0.4) node[text width=60,align=center]{$n\ind{1}$ Layer 1\\ Neurons \\TEM$_i\ind{1}$};

        \draw (\CenterX + \TEMSpikeLineSpan + 0.6*\TEMSpikeLineSpan+ \FilterSpacing + \FilterToTEMLineWidth, \lineStartY -2*\TEMRectHeight -0.2) node[text width=50,align=center] {$n\ind{2}$  outputs\\ $\lbrace t_{i,\ell}\ind{2} \rbrace_\ell$};
        
        \draw (\CenterX - 2*\TEMSpikeLineSpan, \lineStartY -2*\TEMRectHeight -0.4) node[text width=50,align=center] {$n\ind{1}$  spike\\ streams\\ $\lbrace t_{i,\ell}\ind{1} \rbrace_\ell$};

    \draw (\FilterXPos + \FilterWidth/2 , \lineStartY -2*\TEMRectHeight -0.2) node[text width=50,align=center] {Synaptic Filter};
    
    \fi
  
}

\draw[color=red] (-8.8,-3.6) rectangle (-2.2,2.5);

\draw[color=red] (8.2,-3.6) rectangle (-1.8,2.5);

\draw [red, decorate,
    decoration = {brace,
        amplitude=8pt},pen colour = red]   (-8.8,2.65) -- (-2.2,2.65) node[pos=0.5,above=7pt,black]{Layer 1};
        
\draw [red, decorate,
    decoration = {brace,
        amplitude=8pt}, pen colour = red]   (-1.8,2.65) -- (8.2,2.65) node[pos=0.5,above=7pt,black]{Layer 2};
\end{tikzpicture}
\end{center}
    \vspace{-2.2em}
    \caption{A two-layer spiking neural network where each of the individual neurons (marked by circles) is a time encoding machine (TEM) as depicted in Fig.~\ref{fig:TEM circuit}. When learning such an SNN, we assume that we know the synaptic filters as well as the parameters of the TEMs, that we are given a set of examples of input-output pairs and that we wish to learn the weights of the network $\vect{W}\ind{1}$ and $\vect{W}\ind{2}$. One can also consider only the first layer of the SNN and obtain a single-layer SNN, where $x\ind{0}_i(t)$ remains the input but the output neurons are the $\TEMLayerShort{1}$.}
    \label{fig: two layer snn}
    \end{figure*}

Spiking Neural Networks (SNNs) transform their input using nonlinearities that follow simple models of spiking neurons as depicted in Fig.~\ref{fig: two layer snn}. SNNs are gaining in popularity for a number of reasons. They can provide insight on information processing in the brain and can guide experimental studies to validate or refute this insight. They can inspire new learning algorithms for artificial neural networks (ANNs), as SNNs are often made to rely on local operations, thus reducing computational complexity and power load. The hardware needed to implement them can also be very power efficient thanks to the all-or-none nature of SNNs' spiking output.

Despite these motivations, advances in SNNs are quite far from those achieved in the realm of ANNs. While neuromorphic hardware exists~\cite{davies2018loihi, akopyan2015truenorth} and numerous learning algorithms have been developed~\cite{bohte2002error, neftci2019surrogate, comsa2020temporal}, implemented on these chips and tested on various tasks~\cite{cordone2021learning,davies2021advancing},  the tasks that SNNs are trained on are still much simpler than those tackled by ANNs~\cite{comsa2020temporal,wunderlich2021event,ma2021temporal}, because training SNNs on more complex data seems prohibitively difficult. This implies that more progress needs to be made in understanding how to train them.

The difficulty in training SNNs lies in the discontinuity of the function applied by spiking neurons. These neurons fire spikes when their input reaches a threshold---a behavior which results in a discontinuity, posing problems when derivatives are required for backpropagation.
Current approaches to training SNNs avoid this discontinuity in different ways, whether by using spike rates instead of times~\cite{boerlin2013predictive}, using surrogate gradients~\cite{neftci2019surrogate}, or using spike times as a continuous variable with respect to which differentiation is done when performing backpropagation~\cite{bohte2002error, comsa2020temporal, wunderlich2021event}.

An under-explored approach, however, comes from the theory behind time encoding machines (TEMs). This approach can provide a different perspective on learning SNNs which not only avoids the discontinuity mentioned above, but bypasses the backpropagation algorithm altogether. TEMs are devices that filter their input and fire spikes when this filtered version crosses a (potentially time-varying) threshold. These devices can mimic different neuron models, from integrate-and-fire to Hodgkin-Huxley neurons~\cite{gontier2014sampling, lazar2004timerefractory, lazar2008faithful, lazar2010population}.

TEMs can encode and reconstruct spike streams that are passed through various filters~\cite{alexandru2019reconstructing, rudresh2020time,kamath2021time}, such as the alpha synaptic function which is often considered in simpler models of neurons~\cite{hilton2021guaranteed}.
TEMs can also be used to encode and decode weighted sums of input, following an architecture that is reminiscent of a layer of a feedforward network~\cite{adam2020encoding}.
Many of these results are possible because the recovery from time encoding can be formulated as problem of linear constraint satisfaction~\cite{thao2020time}.

In this paper, we start to bridge the gap between TEMs and SNNs by formulating the training problem of SNNs as a  constraint satisfaction problem. This formulation allows us to understand the power of spikes compared to traditional ANN nonlinearities that are graded (i.e. have varying amplitudes) and synchronous.

We assume that we want an SNN to learn to associate certain inputs with corresponding output spike streams, a task which we further assume to be feasible for the particular network used. For a single-layer SNN, when using the TEM formulation, we will see that the SNN's weights can be learnt by solving for one set of linear equations.
We can then build on this result to train two-layer SNNs using the TEM formulation. In the latter scenario, a key ingredient at play is the all-or-none and asynchronous nature of the spikes within an SNN, which will allow layers to be trained one after the other.

The work in this paper is a first attempt at training SNNs using the TEM paradigm, allowing one to bypass traditional backpropagation algorithms, and providing a new tool to tackle this problem.

\section{Background}
\subsection{Time encoding}

\begin{figure}[tb]
	\begin{minipage}[b]{0.85\linewidth}
		\centering
        \begin{tikzpicture}[scale = 1]

    \tikzmath{
        \inputstartx = 0;
        \inputstarty = 0.4;
        \inputlen = 1;
    }
    \draw[->] (\inputstartx, \inputstarty) node[anchor = south west] {$y(t)$} -- (\inputlen, \inputstarty);

    \tikzmath{
        \circleradius = 0.25;
        \circlecenterx = \inputstartx+\inputlen+\circleradius;
        \circlecentery = \inputstarty;
    }
    \draw (\circlecenterx, \circlecentery) circle (\circleradius) node {$+$};

    \tikzmath{
        \biaslinelen = 0.5;
        \biaslinex = \circlecenterx;
        \biasliney = \circlecentery -  \circleradius - \biaslinelen;
    }
    \draw[->] (\biaslinex, \biasliney) node[below] {$\bias$} --  (\biaslinex, \biasliney+ \biaslinelen);

    \tikzmath{
        \tointlinex = \circlecenterx + \circleradius;
        \tointliney = \circlecentery;
        \tointlinelen = 0.5;
    }
    \draw[->] (\tointlinex, \tointliney) -- (\tointlinex + \tointlinelen, \tointliney);

    \tikzmath{
        \integboxheight = 1;
        \integboxwidth = 1.5;
        \integboxx = \tointlinex + \tointlinelen;
        \integboxy = \tointliney - \integboxheight/2;
    }
    
    \draw (\integboxx, \integboxy) rectangle (\integboxx+\integboxwidth, \integboxy+\integboxheight) node[midway] {$\frac{1}{\kappa}\int$};

    \tikzmath{
        \tocomplinex = \integboxx + \integboxwidth;
        \tocompliney = \tointliney;
        \tocomplinelen = 1.5;
    }
    \draw[->] (\tocomplinex, \tocompliney) -- (\tocomplinex + \tocomplinelen, \tocompliney);
    
    \tikzmath{
        \compboxheight = 1;
        \compboxwidth = 1.5;
        \compboxx = \tocomplinex + \tocomplinelen;
        \compboxy = \tocompliney - \compboxheight/2;
    }
    
    \draw (\compboxx, \compboxy) rectangle (\compboxx+\compboxwidth, \compboxy+\compboxheight) node[midway] {\Large $>$};
    
    \tikzmath{
        \deltalinkx = \compboxx;
        \deltalinky = \compboxy + 0.75\compboxheight;
        \deltalinkwidth = 0.5;
        \deltalinkheight = 0.25;
    }
    \draw[->] (\deltalinkx - \deltalinkwidth, \deltalinky + \deltalinkheight) -- (\deltalinkx- \deltalinkwidth, \deltalinky) -- (\deltalinkx, \deltalinky);
    
    \tikzmath{
        \deltaboxwidth = 0.5;
        \deltaboxheight = 0.5;
        \deltaboxx = \deltalinkx - \deltalinkwidth - \deltaboxwidth/2;
        \deltaboxy = \deltalinky + \deltalinkheight ;
    }
    
    \draw (\deltaboxx, \deltaboxy) rectangle  node {$\delta$} (\deltaboxx + \deltaboxwidth, \deltaboxy+ \deltaboxheight);
    
    \tikzmath{
        \tooutlinex = \compboxx + \compboxwidth;
        \tooutliney = \tocompliney;
        \tooutlinelen = 1.5;
    }
    \draw[->] (\tooutlinex, \tooutliney) node[anchor = south west] {$t_k$} -- (\tooutlinex + \tooutlinelen, \tooutliney);
    
    \tikzmath{
        \feedbacklinex = \tooutlinex +0.5;
        \feedbackliney = \tooutliney;
        \feedbacklineheight = 1;
        \feedbacklineendy = \integboxy;
        \feedbacklineendx = \integboxx + \integboxwidth/2;
    }
    \draw[dashed, ->] (\feedbacklinex, \feedbackliney) -- (\feedbacklinex , \feedbackliney- \feedbacklineheight) -- node[below] {\textit{Spike triggered reset}} (\feedbacklineendx , \feedbackliney - \feedbacklineheight) --    (\feedbacklineendx, \feedbacklineendy);
\end{tikzpicture}
	\end{minipage}
	\vspace{-2em}
	\caption{Circuit of a Time Encoding Machine, with input $y(t)$, threshold $\delta$, integrator constant $\kappa$ and bias $\bias$.}
	\label{fig:TEM circuit}
	\end{figure}
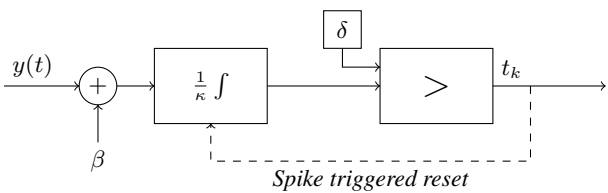

In this paper, we consider integrate-and-fire time encoding machines (TEMs), as depicted in Fig.~\ref{fig:TEM circuit}. 

    Such a TEM with integrator constant $\kappa$, threshold $\delta$, and bias $\bias$ takes an input signal $y(t)$, adds $\bias$ to it and integrates the result, scaled by $1/\kappa$, until $\delta$ is reached. Once this threshold is reached, the time $t_\ell$ at which it is reached is recorded, the value of the integrator resets to $-\delta$ and the mechanism restarts. We assume that the machine emits a Dirac delta at the integrator reset, and we call the Dirac time $t_\ell$ a \textit{spike time}. \todo[inline]{need to decide if i mention spiking business}
Nyquist-like results have been established about the recovery of bandlimited signals using single TEMs~\cite{lazar2003time, gontier2014sampling} as well as populations of TEMs~\cite{lazar2008faithful, lazar2010population,adam2020sampling, adam2021asynchrony}. The results mainly follow from the realization that the spike times $\lbrace t_\ell \rbrace$ of a TEM provide linear constraints on the input $y(t)$ of the machine:
\begin{equation}
\label{eq: times to integrals}
    \int_{t_\ell}^{t_{\ell+1}} y(u)\, du = 2\kappa\delta - \bias(t_{\ell+1}-t_\ell).
\end{equation}

On the other hand, results have also targeted input signals with finite rate of innovation (FRI)~\cite{vetterli2002sampling}. We mention, for example, the work in~\cite{alexandru2019reconstructing} which uses exponential and polynomial splines to recover streams of Diracs (or spikes), and the work in~\cite{hilton2021guaranteed} which recovers streams of Diracs convolved with alpha-synaptic kernels.

Next, we summarize findings from~\cite{rudresh2020time, kamath2021time} about time encoding streams of Diracs, to lay the foundations for the results in this paper.

\subsection{Time encoding of finite rate of innovation signals}
\label{sec: fri-tem}
Consider streams of weighted Dirac deltas periodized with period $T$:
\begin{equation}
\label{eq: dirac stream}
    y(t) = \sum_{N\in \mathbb{Z}} \sum_{k=1}^K c_k \delta(t-t_k - NT).
\end{equation}

Assume the amplitudes $c_k$ and Dirac times $t_k$ are unknown, it is possible to recover them by time encoding a filtered version of $y(t)$.

First, we say that a filter $h(t)$ satisfies the alias cancellation property, if its Fourier transform $H(\omega)$ satisfies, for $\omega_0 = 2\pi/T$,
\begin{align}
\label{eq: alias cancellation}
    H(k\omega_0) = \left\{ 
    \begin{array}{ll}
    c_k \neq 0 & \textrm{if} \,\, k=\lbrace -K, \cdots, K\rbrace,\\ 
    0 & \textrm{otherwise.}
    \end{array}
    \right.
\end{align}

Assume a signal as in~\eqref{eq: dirac stream} is passed through a filter that satisfies the alias cancellation property as in~\eqref{eq: alias cancellation} and is then sampled using a TEM. Lemma 4 in~\cite{kamath2021time} indicates that the Fourier series coefficients $Y(k\omega_0),\, k = -K, \cdots, K $ of the input can then be recovered from time encoding if more than $2K+2$ spikes are emitted. 
 
Consequently, given that $2K+1$ Fourier series coefficients of the input can be recovered, one can find an annihilating filter for it~\cite{vetterli2002sampling}, the roots $u_k$ of which  will provide information about the timing of the Diracs:
    $u_k = \exp \left(-\mathbf{j}2\pi t_k/T\right).$
The Dirac amplitudes $c_k$ are then estimated  by simply solving a least squares problem.

Having summarized key technical results needed for this paper, we now move on to set the scene of training of SNNs.

\subsection{Learning rules for SNN training}

There is ongoing debate about the modality the brain uses to encode information: is it spike times or spike rates? Of course, this debate also extends to SNNs, where the question targets the quantity used to perform learning.

Evidence has been found for both spike rate and spike time coding in the brain, and the preference in coding scheme depends on the area in question. It is argued, for example, that some tasks are performed too quickly for them to depend on the computation of spike rates and rather rely on quantities such as the time to first spike~\cite{vanrullen2005spike}.

Following these two schools of thought when it comes to brain activity, we also see two trends for learning rules of SNNs.

In the spike rate paradigm, a neuron is often assumed to emit spikes according to a Poisson process with a varying and information-carrying rate (which is its input) and operations are done by calculating and using the rate of the neuron over moving windows as done e.g. in~\cite{boerlin2013predictive}.

In the spike timing paradigm, there exist different training approaches to SNNs. For example, single-layered SNNs can be learned using \textit{spike timing dependent plasticity}~\cite{pfister2006triplets}. For learning rules for deeper SNNs, we point to the SpikeProp algorithm which computes gradients with respect to the timing of spikes~\cite{bohte2002error}.
A similar line of work was also explored by Comsa \emph{ et al.}~\cite{comsa2020temporal} and Wunderlich \emph{et al.}~\cite{wunderlich2021event}, often prioritizing quantities such as the \textit{time to first spike}, i.e. the timing of the first spike of an output neuron.

We will next see how results from time encoding can inspire learning algorithms of a different kind.


\section{Learning a Single-Layer SNN}
\label{sec: one layer}

\subsection{Problem setup}

We would like to understand how to make an SNN learn a desired behavior, i.e. to learn to generate, for a set of inputs,  corresponding output spike streams. 

To begin tackling this problem, we first consider the case of a single-layered SNN, i.e. we focus on the first layer in Fig.~\ref{fig: two layer snn}. The SNN has nonlinearities which are modeled by integrate-and-fire neurons, i.e. TEMs as depicted in Fig.~\ref{fig:TEM circuit}, and are assumed to have a known set of parameters $\kappa$, $\delta$ and $\bias$. 

We would like to recover an unknown weight matrix $\vect{W}\ind{1}$ and to do so, we are given a set of examples. Each of these examples is composed of an input $\vect{x}(t) = \left[ x_1\ind{0}(t), \cdots,  x_{n\ind{0}}\ind{0}(t)\right]^T$ which is a collection of $n\ind{0}$ signals, and of the corresponding target output spike streams $\lbrace t_{i,\ell}\ind{1} \rbrace_\ell$ for each of the $n\ind{1}$ neurons TEM$_i\ind{1}$.

Given the known input and output of each example, we would like to find the weights $\vect{W}\ind{1}$ such that the SNN generates the correct output for each input of the examples.

\subsection{From spikes to linear constraints}

The recovery of the weights $\vect{W}\ind{1}$ can be performed in a row by row fashion.
In fact, the timing of the spikes $\lbrace t_{i,\ell}\ind{1} \rbrace_\ell$ of TEM$_i\ind{1}$ provide linear constraints on the input $\sum_j w_{i,j}x_j\ind{0}(t)$ to TEM$_i\ind{1}$ and therefore on the weight matrix row $w_{i,:}$

\begin{equation}
    \label{eq: constraint on weights}
    b_\ell\ind{i} = \int_{t_{i,\ell}\ind{1}}^{t_{i,\ell+1}\ind{1}} \sum_j w_{i,j}x_j\ind{0}(t) =
    \sum_j w_{i,j} \int_{t_{i,\ell}\ind{1}}^{t_{i,\ell+1}\ind{1}} x_j\ind{0}(t),
\end{equation}
where $b_\ell\ind{i} = 2\kappa\delta - \bias(t_{i,\ell}\ind{1}-t_{i,\ell+1}\ind{1})$ as mentioned in~\eqref{eq: times to integrals}.

Given that the inputs $\vect{x}(t)$ are known, their integrals needed for~\eqref{eq: constraint on weights} are also known. It is also possible to further show that the spike times will almost surely (a.s.) provide constraints that are linearly independent, if one assumes, for instance, that the $\vect{x}(t)$ are periodic bandlimited signals and have their coefficients drawn from a Lipschitz-continuous probability distribution~\cite{pacholska2020matrix}.

Notice that, in~\eqref{eq: constraint on weights}, we assume we have one example with corresponding input and output spike stream, but this can be extended to include multiple examples, by concatenating the constraints from each example on $\vect{W}\ind{1}$ into a single measurement matrix and vector.
\subsection{Simulations}

\begin{figure}[tb]
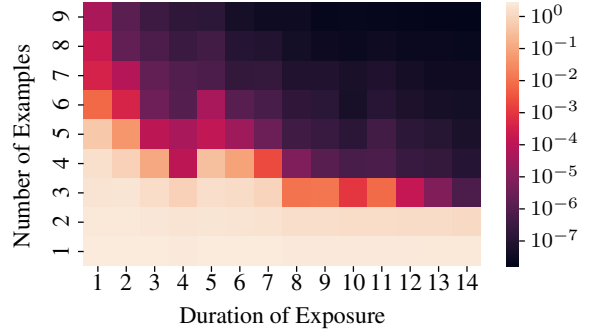

    \centering
    \def\svgwidth{0.9\columnwidth}
    \subfile{../Figures/single_layer_spikes_vs_examples.tex}
    \vspace{-1em}
    \caption{    \label{fig: Simulations single layer - figure 1}
Heatmap of the reconstruction error of the weight matrix $\hat{\vect{W}}\ind{1}$, using the $L^2$ norm, as a function of the number of examples and the exposure time used to train the network.}
\end{figure}

\begin{figure}[tb]
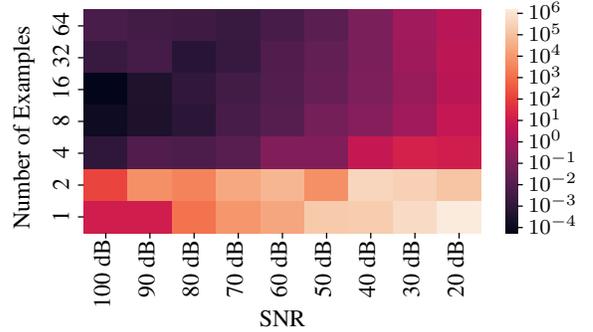

    \centering
    \def\svgwidth{0.9\columnwidth}
    \subfile{../Figures/single_layer_examples_vs_noise.tex}
        \vspace{-1em}
    \caption{    \label{fig: Simulations single layer - figure 2}
    Heatmap of the reconstruction error of the weight matrix $\hat{\vect{W}}\ind{1}$, using the $L^2$ norm, as a function of the number of examples and the SNR of the spike times used to learn the network.}
\end{figure}

We assume that we would like to train an SNN that has $n\ind{0} = 20$ inputs and $n\ind{1} = 5$ outputs, using a varying number of examples.

To do so, we create a \textit{Simulator SNN} with weights $\vect{W}\ind{1}$ drawn uniformly at random and process a set of input signals through the Simulator SNN to obtain our \textit{ground truth} output spike streams. The input signals are here assumed to be periodic bandlimited, with Fourier series coefficients drawn uniformly at random.

Then, we learn the weights $\hat{\vect{W}}\ind{1}$ of a \textit{Learned SNN} that match these examples, by finding the least squares solution to~\eqref{eq: constraint on weights}, which can either be done in one shot or through gradient descent.

Note that the Learned SNN has the exact same architecture as the Simulator SNN, and only the weights $\hat{\vect{W}}\ind{1}$ are being learned, so that we are essentially simulating a teacher-student setup where teacher and student are single-layer SNNs.

To evaluate the success of our learning, we look at the mean squared error of our estimate of the weights $\hat{\vect{W}}\ind{1}$. This is not the standard evaluation metric when training neural networks, but it is a pertinent measure here as the weights should be recoverable exactly.

We study the reconstruction error on the weights after training using a varying number of examples and varying exposure duration per example, in Fig.~\ref{fig: Simulations single layer - figure 1}. By exposure duration, we mean the time over which the Simulator SNN is exposed to the input and allowed to spike. The higher the exposure duration, the more spikes (and constraints) are generated.

Then, in Fig.~\ref{fig: Simulations single layer - figure 2}, we provide simulations where we introduce uniform noise on the spike times, varying the signal-to-noise ratio (SNR) of the spike times, and evaluate the reconstruction error on the weights after learning. Notice how increasing the number of examples can partly compensate for higher noise levels.

\section{Learning a Two-Layer SNN}
\subsection{Problem formulation}

We have seen that we can train a single-layer SNN in a one-shot inversion of linear equations~\eqref{eq: constraint on weights}, and now consider the case where we have two layers of neurons or TEMs as depicted by Fig.~\ref{fig: two layer snn}.

For a certain input $\vect{x}(t)$ to the network, this input is passed through a weight matrix $\vect{W}\ind{1}$, through a layer of nonlinearities $\TEMLayer{1}$, that output streams of spikes $\lbrace t_{i,\ell}\ind{1} \rbrace_\ell$, which are passed through a second weight matrix $\vect{W}\ind{2}$, through a filter (which can be likened to synaptic filters in real neurons), and then through the second layer of nonlinearities $\mathrm{TEM}_i\ind{2},$ ${i=1,\cdots, n\ind{2}}$, each of which outputs a stream of spikes $\lbrace t_{i,\ell}\ind{2} \rbrace_\ell$.

As before, we assume that we would like to train the network, i.e. find the weight matrices $\vect{W}\ind{1}$ and $\vect{W}\ind{2}$. To do so, we have a set of examples, each of which is composed of an input $\vect{x}(t)$---a collection of $n\ind{0}$ signals $x_i\ind{0}(t)$---and of the corresponding target output spike streams $\lbrace t_{i,\ell}\ind{2} \rbrace_\ell$ for each of the $n\ind{2}$ output neurons $\TEMLayer{2}$.
We further assume that the filters that are used between the first and second layer of TEMs satisfy the alias cancellation property as defined in Section~\ref{sec: fri-tem}.

Following a similar approach to the one in Section~\ref{sec: one layer}, we use the examples to constrain the weight matrices $\vect{W}\ind{1}$ and $\vect{W}\ind{2}$. 
Of course, this does not result in \textit{linear} constraints on $\vect{W}\ind{1}$ and $\vect{W}\ind{2}$. Therefore, we propose an approach which performs a layer by layer recovery of the weight matrices, starting with $\vect{W}\ind{2}$.

\begin{figure}[tb]
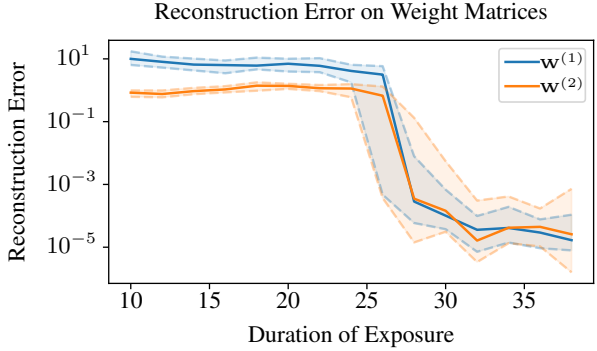

    \centering
    \def\svgwidth{0.92\columnwidth}
    \subfile{../Figures/two_layers_sim.tex}
    \vspace{-0.5em}
    \caption{Reconstruction Error, using the $L^2$ norm, of $\vect{W}\ind{1}$ and $\vect{W}\ind{2}$ of a two-layer SNN, as the duration of exposure increases. We plot the median and the first and second quartiles of the error for 50 randomly generated networks, each trained with two examples.}
    \label{fig: simulation two layers}
\end{figure}
\subsection{Proposed solution}
\label{sec: two layer solution}
Learning both weight matrices $\vect{W}\ind{1}$ and $\vect{W}\ind{2}$ would be straightforward if the output of the hidden layer $\TEMLayerShort{1}$ were known. As this is not the case, we look to recover it. The trick we use lies in the all-or-none nature of the spikes output by the $\TEMLayerShort{1}$. These outputs are streams of Diracs, as described in~\eqref{eq: dirac stream}, but with $c_k = 1$, given that a TEM encodes its input using spike times only.

Therefore, the inputs to the second layer $\TEMLayerShort{2}$ will also be streams of Diracs, the \textit{times} of which are dictated by the outputs of $\TEMLayerShort{1}$ and the \textit{weights} of which are dictated by $\vect{W}\ind{2}$.

Consequently, if the $\TEMLayerShort{2}$ spikes often enough (and we will clarify this shortly), the Fourier series coefficients of the inputs to the $\TEMLayerShort{2}$ can be recovered, and a joint annihilating filter can be found for all of these inputs. It is then easy to recover the locations of the Diracs, then assign them to different TEMs of $\TEMLayerShort{1}$ while recovering $\vect{W}\ind{2}$ (up to a permutation) simply by using a k-means algorithm. Note that this is only possible because of the all-or-none and asynchronous nature of the spikes. This is not possible with classical nonlinearities that are graded and simultaneous. Once the outputs of $\TEMLayerShort{1}$ are known, it is easy to recover the weights $\vect{W}\ind{1}$ as we did in Section~\ref{sec: one layer}. 

Let us now specify how many spikes at the output of $\TEMLayerShort{2}$ are needed in order for the output of $\TEMLayerShort{1}$ and for $\vect{W}\ind{2}$ to be recoverable. Following the result in~\cite{kamath2021time} and reformulated in~\ref{sec: fri-tem}, we need every TEM at the output to generate $(2L+2)n\ind{1}$ spikes if each of the TEMs in the hidden layer fires $L$ spikes. We will discuss the implications of this in the conclusion.

\subsection{Simulations}

To provide a proof a concept, we design simulations where we again work with a Simulator SNN and a Learned SNN. Both SNNs are assumed to have $n\ind{0}=n\ind{1} = 2$ input and hidden nodes and $n\ind{2} = 4$ output nodes.

The Simulator SNN has its weight matrices $\vect{W}\ind{1}$ and $\vect{W}\ind{2}$ drawn uniformly at random and is used to obtain the spike stream outputs for the inputs of (in this case) two examples. 
Then, the Learned SNN ``learns'' the appropriate weight matrices $\hat{\vect{W}}\ind{1}$ and $\hat{\vect{W}}\ind{2}$ to reproduce the examples using our described approach.

We show the reconstruction error of the weight matrices as the duration of exposure increases. As before, the longer the duration of the exposure, the more spikes the SNN can generate at the output.

\section{Conclusion}

We have suggested an approach that learns the weights of SNNs in a layer-by-layer approach, leveraging the all-or-none and asynchronous nature of the spikes. In fact, these properties of spikes allow the recovery of weight matrices up to permutations, whereas a similar approach applied to ANNs, for example, would never allow recovery beyond Hermitian transformations.

It is clear, however, that there remains sub-problems to be solved for this approach to be actionable in reality. First, our approach currently only deals with the case where the desired task is perfectly learnable.
Second, we currently assume that we know what output spike stream to associate with input examples, which is not applicable in every situation. Such a setup can be conceived when training denoising autoencoders, for example, but it should be clarified how one can translate generic loss functions to corresponding spike streams that minimize them.

Moreover, while our approach can technically be extended to more layers, we currently require that the number of spikes at the output of any layer scale as the \textit{total} number of spikes in the previous layer, \textit{multiplied} by the number of nodes in the given layer. As a result, the number of spikes required scales badly with the number of layers. We hope that this requirement can be optimized by leveraging the joint annihilation property of the outputs of each layer.

Finally, note that our results assume that the output spike stream of $\TEMLayerShort{1}$ is periodic, which we can enforce in our simulations but is, in reality, more difficult to ensure.

While these questions remain unsolved, they motivate different research directions, each of which is worthy of attention for its own sake.

\bibliographystyle{IEEEbib}

\bibliography{main}

\begin{thebibliography}{10}

\bibitem{davies2018loihi}
Mike Davies, Narayan Srinivasa, Tsung-Han Lin, Gautham Chinya, Yongqiang Cao,
  Sri~Harsha Choday, Georgios Dimou, Prasad Joshi, Nabil Imam, Shweta Jain,
  et~al.,
\newblock ``Loihi: A neuromorphic manycore processor with on-chip learning,''
\newblock {\em Ieee Micro}, vol. 38, no. 1, pp. 82--99, 2018.

\bibitem{akopyan2015truenorth}
Filipp Akopyan, Jun Sawada, Andrew Cassidy, Rodrigo Alvarez-Icaza, John Arthur,
  Paul Merolla, Nabil Imam, Yutaka Nakamura, Pallab Datta, Gi-Joon Nam, et~al.,
\newblock ``Truenorth: Design and tool flow of a 65 mw 1 million neuron
  programmable neurosynaptic chip,''
\newblock {\em IEEE transactions on computer-aided design of integrated
  circuits and systems}, vol. 34, no. 10, pp. 1537--1557, 2015.

\bibitem{bohte2002error}
Sander~M Bohte, Joost~N Kok, and Han La~Poutre,
\newblock ``Error-backpropagation in temporally encoded networks of spiking
  neurons,''
\newblock {\em Neurocomputing}, vol. 48, no. 1-4, pp. 17--37, 2002.

\bibitem{neftci2019surrogate}
Emre~O Neftci, Hesham Mostafa, and Friedemann Zenke,
\newblock ``Surrogate gradient learning in spiking neural networks: Bringing
  the power of gradient-based optimization to spiking neural networks,''
\newblock {\em IEEE Signal Processing Magazine}, vol. 36, no. 6, pp. 51--63,
  2019.

\bibitem{comsa2020temporal}
Iulia~M Comsa, Krzysztof Potempa, Luca Versari, Thomas Fischbacher, Andrea
  Gesmundo, and Jyrki Alakuijala,
\newblock ``Temporal coding in spiking neural networks with alpha synaptic
  function,''
\newblock in {\em ICASSP 2020-2020 IEEE International Conference on Acoustics,
  Speech and Signal Processing (ICASSP)}. IEEE, 2020, pp. 8529--8533.

\bibitem{cordone2021learning}
Lo{\"\i}c Cordone, Beno{\^\i}t Miramond, and Sonia Ferrante,
\newblock ``Learning from event cameras with sparse spiking convolutional
  neural networks,''
\newblock {\em arXiv preprint arXiv:2104.12579}, 2021.

\bibitem{davies2021advancing}
Mike Davies, Andreas Wild, Garrick Orchard, Yulia Sandamirskaya, Gabriel
  A~Fonseca Guerra, Prasad Joshi, Philipp Plank, and Sumedh~R Risbud,
\newblock ``Advancing neuromorphic computing with loihi: A survey of results
  and outlook,''
\newblock {\em Proceedings of the IEEE}, vol. 109, no. 5, pp. 911--934, 2021.

\bibitem{wunderlich2021event}
Timo~C Wunderlich and Christian Pehle,
\newblock ``Event-based backpropagation can compute exact gradients for spiking
  neural networks,''
\newblock {\em Scientific Reports}, vol. 11, no. 1, pp. 1--17, 2021.

\bibitem{ma2021temporal}
Chenxiang Ma, Junhai Xu, and Qiang Yu,
\newblock ``Temporal dependent local learning for deep spiking neural
  networks,''
\newblock in {\em 2021 International Joint Conference on Neural Networks
  (IJCNN)}. IEEE, 2021, pp. 1--7.

\bibitem{boerlin2013predictive}
Martin Boerlin, Christian~K Machens, and Sophie Den{\`e}ve,
\newblock ``Predictive coding of dynamical variables in balanced spiking
  networks,''
\newblock {\em PLoS computational biology}, vol. 9, no. 11, pp. e1003258, 2013.

\bibitem{gontier2014sampling}
David Gontier and Martin Vetterli,
\newblock ``Sampling based on timing: Time encoding machines on shift-invariant
  subspaces,''
\newblock {\em Applied and Computational Harmonic Analysis}, vol. 36, no. 1,
  pp. 63--78, 2014.

\bibitem{lazar2004timerefractory}
Aurel~A Lazar,
\newblock ``Time encoding with an integrate-and-fire neuron with a refractory
  period,''
\newblock {\em Neurocomputing}, vol. 58, pp. 53--58, 2004.

\bibitem{lazar2008faithful}
Aurel~A Lazar and Eftychios~A Pnevmatikakis,
\newblock ``Faithful representation of stimuli with a population of
  integrate-and-fire neurons,''
\newblock {\em Neural Computation}, vol. 20, no. 11, pp. 2715--2744, 2008.

\bibitem{lazar2010population}
Aurel~A Lazar,
\newblock ``Population encoding with {H}odgkin--{H}uxley neurons,''
\newblock {\em IEEE Transactions on Information Theory/Professional Technical
  Group on Information Theory}, vol. 56, no. 2, 2010.

\bibitem{alexandru2019reconstructing}
Roxana Alexandru and Pier~Luigi Dragotti,
\newblock ``Reconstructing classes of non-bandlimited signals from time encoded
  information,''
\newblock {\em IEEE Transactions on Signal Processing}, vol. 68, pp. 747--763,
  2019.

\bibitem{rudresh2020time}
Sunil Rudresh, Abijith~Jagannath Kamath, and Chandra~Sekhar Seelamantula,
\newblock ``A time-based sampling framework for finite-rate-of-innovation
  signals,''
\newblock in {\em ICASSP 2020-2020 IEEE International Conference on Acoustics,
  Speech and Signal Processing (ICASSP)}. IEEE, 2020, pp. 5585--5589.

\bibitem{kamath2021time}
Abijith~Jagannath Kamath, Sunil Rudresh, and Chandra~Sekhar Seelamantula,
\newblock ``Time encoding of finite-rate-of-innovation signals,''
\newblock {\em arXiv preprint arXiv:2107.03344}, 2021.

\bibitem{hilton2021guaranteed}
Marek Hilton, Roxana Alexandru, and Pier~Luigi Dragotti,
\newblock ``Guaranteed reconstruction from integrate-and-fire neurons with
  alpha synaptic activation,''
\newblock in {\em ICASSP 2021-2021 IEEE International Conference on Acoustics,
  Speech and Signal Processing (ICASSP)}. IEEE, 2021, pp. 5474--5478.

\bibitem{adam2020encoding}
Karen Adam, Adam Scholefield, and Martin Vetterli,
\newblock ``Encoding and decoding mixed bandlimited signals using spiking
  integrate-and-fire neurons,''
\newblock in {\em ICASSP 2020-2020 IEEE International Conference on Acoustics,
  Speech and Signal Processing (ICASSP)}. IEEE, 2020, pp. 9264--9268.

\bibitem{thao2020time}
Nguyen~T Thao and Dominik Rzepka,
\newblock ``Time encoding of bandlimited signals: reconstruction by
  pseudo-inversion and time-varying multiplierless fir filtering,''
\newblock {\em IEEE Transactions on Signal Processing}, vol. 69, pp. 341--356,
  2020.

\bibitem{lazar2003time}
Aurel~A Lazar and L{\'a}szl{\'o}~T T{\'o}th,
\newblock ``Time encoding and perfect recovery of bandlimited signals,''
\newblock in {\em 2003 IEEE International Conference on Acoustics, Speech, and
  Signal Processing, 2003. Proceedings.(ICASSP'03).} IEEE, 2003, vol.~6, pp.
  VI--709.

\bibitem{adam2020sampling}
Karen Adam, Adam Scholefield, and Martin Vetterli,
\newblock ``Sampling and reconstruction of bandlimited signals with
  multi-channel time encoding,''
\newblock {\em IEEE Transactions on Signal Processing}, vol. 68, pp.
  1105--1119, 2020.

\bibitem{adam2021asynchrony}
Karen Adam, Adam Scholefield, and Martin Vetterli,
\newblock ``Asynchrony increases efficiency: Time encoding of videos and
  low-rank signals,''
\newblock {\em arXiv preprint arXiv:2104.14511}, 2021.

\bibitem{vetterli2002sampling}
Martin Vetterli, Pina Marziliano, and Thierry Blu,
\newblock ``Sampling signals with finite rate of innovation,''
\newblock {\em IEEE transactions on Signal Processing}, vol. 50, no. 6, pp.
  1417--1428, 2002.

\bibitem{vanrullen2005spike}
Rufin VanRullen, Rudy Guyonneau, and Simon~J Thorpe,
\newblock ``Spike times make sense,''
\newblock {\em Trends in neurosciences}, vol. 28, no. 1, pp. 1--4, 2005.

\bibitem{pfister2006triplets}
Jean-Pascal Pfister and Wulfram Gerstner,
\newblock ``Triplets of spikes in a model of spike timing-dependent
  plasticity,''
\newblock {\em Journal of Neuroscience}, vol. 26, no. 38, pp. 9673--9682, 2006.

\bibitem{pacholska2020matrix}
Michalina Pacholska, Karen Adam, Adam Scholefield, and Martin Vetterli,
\newblock ``Matrix recovery from bilinear and quadratic measurements,''
\newblock {\em arXiv preprint arXiv:2001.04933}, 2020.

\end{thebibliography}
\end{document}